\documentclass[10pt,twocolumn,letterpaper]{article}

\pdfoutput=1

\usepackage{iccv}
\usepackage{times}
\usepackage{epsfig}
\usepackage{graphicx}
\usepackage{amsmath}
\usepackage{amssymb}
\usepackage[super]{nth}

\usepackage{booktabs}
\usepackage[normalem]{ulem}
\useunder{\uline}{\ul}{}
\usepackage{longtable}

\usepackage{subfig}

\usepackage[accsupp]{axessibility}

\usepackage[pagebackref=true,breaklinks=true,letterpaper=true,colorlinks,bookmarks=false]{hyperref}

\iccvfinalcopy 

\def\iccvPaperID{7726}

\ificcvfinal\fi

\DeclareRobustCommand{\myparagraph}[1]
{\noindent\textbf{#1}}

\newcommand\blfootnote[1]{%
  \begingroup
  \renewcommand\thefootnote{}\footnote{#1}%
  \addtocounter{footnote}{-1}%
  \endgroup
}

\begin{document}

\title{Space-Time-Separable Graph Convolutional Network for Pose Forecasting}

\author{Theodoros Sofianos$^{\dag}$, Alessio Sampieri$^{\dag}$, Luca Franco and Fabio Galasso\\
Sapienza University of Rome, Italy\\

}

\maketitle
\ificcvfinal\thispagestyle{empty}\fi

\begin{abstract}

Human pose forecasting is a complex structured-data sequence-modelling task, which has received increasing attention, also due to numerous potential applications. Research has mainly addressed the temporal dimension as time series and the interaction of human body joints with a kinematic tree or by a graph. This has decoupled the two aspects and leveraged progress from the relevant fields, but it has also limited the understanding of the complex structural joint spatio-temporal dynamics of the human pose.
Here we propose a novel Space-Time-Separable Graph Convolutional Network (STS-GCN) for pose forecasting. For the first time, STS-GCN models the human pose dynamics only with a graph convolutional network (GCN), including the temporal evolution and the spatial joint interaction within a single-graph framework, which allows the cross-talk of motion and spatial correlations. Concurrently, STS-GCN is the first space-time-separable GCN: the space-time graph connectivity is factored into space and time affinity matrices, which bottlenecks the space-time cross-talk, while enabling full joint-joint and time-time correlations. Both affinity matrices are learnt end-to-end, which results in connections substantially deviating from the standard kinematic tree and the linear-time time series.
In experimental evaluation on three complex, recent and large-scale benchmarks, Human3.6M~\cite{Human36M}, AMASS~\cite{AMASS} and 3DPW~\cite{3DPW}, STS-GCN outperforms the state-of-the-art, surpassing the current best technique~\cite{Mao20DCT} by over 32\% in average at the most difficult long-term predictions, while only requiring 
1.7\% of its parameters. We explain the results qualitatively and illustrate the graph interactions by the factored joint-joint and time-time learnt graph connections.

Our source code is available at:\\ \url{https://github.com/FraLuca/STSGCN}

\blfootnote{ $^{\dag}$  indicates equal contribution}

\end{abstract}

\section{Introduction}

\begin{figure}[t!]
\begin{center}
\includegraphics[width=0.5\textwidth]{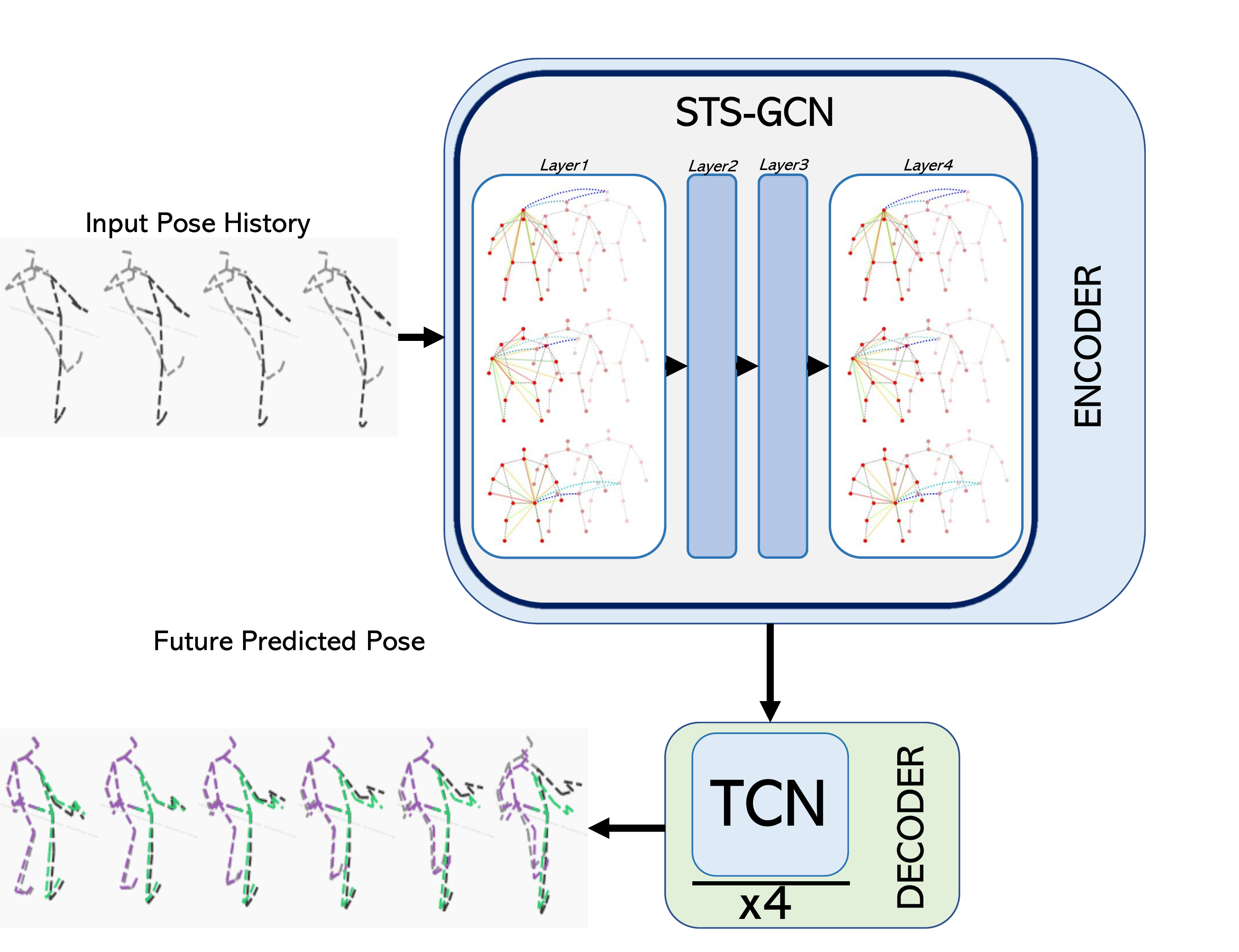}
\end{center}
\caption{\textbf{Overview of the proposed pipeline}. Given a sequence of observed 3D poses, the novel STS-GCN encodes the spatio-temporal body dynamics. The encoded representation serves to predict future poses by means of a Temporal Convolutional Network (TCN).
STS-GCN allows the spatial and temporal interaction of joints, cf.\ green-orange linkage on the Vitruvian man and dashed blue lines connecting joints in time, which are both learnt. But it bottlenecks their cross-talk by a new GCN design with factored space-time adjacency matrices. (Vector image: please zoom in.)}
\label{fig:model_pipeline}
\end{figure}

Forecasting future human poses is the task of modelling the complex structured-sequence of joint spatio-temporal dynamics of the human body. This has received increasing attention due to its manifold applications to autonomous driving~\cite{Paden16}, healthcare~\cite{Troje02}, teleoperations~\cite{Rubagotti19} and collaborative robots~\cite{Koppula13, Unhelkar18}, where e.g.\ anticipating the human motion avoids crashes and helps the robots plan the future.

Research has so far addressed modelling space and time in separate frameworks. Time has generally been modelled with convolutions in the temporal dimension~\cite{Cao20}, with recurrent neural networks (RNN~\cite{Mao20DCT,Mao19LTD,Wang19,Chiu19}, GRU~\cite{Yuan20,Adeli20} and LSTM~\cite{Yuan19}) or with Transformer Networks~\cite{Cai20}.
Space and the interaction of joints has instead been recently modelled by Graph Convolutional Networks (GCN)~\cite{Mao20DCT}, mostly connecting body joints along a kinematic tree. The separate approach has side-stepped the complexity of a joint model across the spatial and temporal dimensions, which are diverse in nature, and has leveraged progress in the relevant fields. However this has also limited the understanding of the complex human body dynamics.

Here we propose to forecast human motion with a novel Space-Time-Separable Graph Convolutional Network (STS-GCN). STS-GCN encodes both the spatial joint-joint and the temporal time-time correlations with a joint spatio-temporal GCN~\cite{Kipf17gcn}. The single-graph framework favors the cross-talk of the body joint interactions and their temporal motion patterns. Further to better performance, using the GCN-only model results in considerably less parameters.

To the best of our knowledge, STS-GCN is the first space-time separable GCN. We realize this by factorizing the graph adjacency matrix $A^{st}$ into $A^{s} A^{t}$. Our intuition is that bottleneck'ing the cross-talk of the spatial joints and the temporal frames helps to improve the interplay of spatial joints and temporal patterns. This differs substantially from recent work~\cite{LaiRejected2018depthwise,Balcilar20depthwisespectral} which separate the graph interactions from the channel convolutions, being therefore depthwise separable. Still both separable designs are advantageous for the reduction of model parameters.

Fig.~\ref{fig:model_pipeline} illustrates the encoder-decoder design of our model. Following the body motion encoding by the STS-GCN, the future pose coordinates are forecast with few simple convolutional layers, generally termed Temporal Convolutional Network (TCN)~\cite{Gehring17,Bai18TCN,Luo18}, robust and fast to train.

Note from Fig.~\ref{fig:model_pipeline} that the factored $A^{s} A^{t}$ graph adjacency matrices are learnt. This results in better performance and it allows us to interpret the joint-joint and the time-time interactions, as we further illustrate in Fig.~\ref{fig:spatial_edges} and in Sec.~\ref{sec:exp}.

In extensive experiments over the modern, challenging and large-scale datasets of Human3.6M~\cite{Human36M}, AMASS~\cite{AMASS} and 3DPW~\cite{3DPW}, we demonstrate that STS-GCN improves over the state-of-the-art. Notably, STS-GCN outperforms the current best technique~\cite{Mao20DCT} by over 32\% on all three datasets, in average at the most difficult long-term predictions, while only adopting 1.7\% of its parameters.

We summarize our main contributions as follows:
\begin{itemize}
    \item We propose the first space-time separable graph convolutional network, which is first to factorize the graph adjacency matrix, rather than depthwise~\cite{LaiRejected2018depthwise,Balcilar20depthwisespectral};
    \item Our space-time human body representation is the first to exclusively use a GCN and it adopts only 1.7\% parameters of the current best competing technique~\cite{Mao20DCT};
    \item We improve on the state-of-the-art by over 32\% on Human3.6M~\cite{Human36M}, AMASS~\cite{AMASS} and 3DPW~\cite{3DPW}, in average at the most challenging long-term predictions;
    \item The joint-joint and time-time graph edge weights are learnt, which allows to explain their interactions.
    
\end{itemize}

\section{Related Work}

Human pose forecasting is a long-standing problem~\cite{Cao20}. We discuss related work by distinguishing the temporal aspects of sequence modelling and the spatial representations. Finally we relate to separable convolutional networks.

\myparagraph{Temporal modelling}
Most recent work in human pose forecasting has leveraged Recurrent Neural Networks (RNN)~\cite{Fragkiadaki15,Jain16,Julieta17,Chiu19,Mao20DCT,Mao19LTD,Wang19}, as well as recurrent variants such as Gated Recurrent Units (GRU)~\cite{Yuan20,Adeli20} and Long Short-Term Memory Networks~\cite{Yuan19}. These techniques are flexible, but they have issues with long-term predictions such as inefficient training and poor long-term memory~\cite{Barsoum17,Li18,Mao19LTD,Mao20DCT}. Research has attempted to tackle this, e.g.\ by training with generative adversarial networks~\cite{Gui18} and by imitation learning~\cite{Tang18,Wang19}.
Emerging trends have adopted (self-)attention to model time~\cite{Mao20DCT,Cai20}, which also applies to model spatial relations~\cite{Tang18,Cai20}.\\
State-of-the-art performance is also attained with convolutional layers~\cite{Butepage17,Li18,Mao19LTD,Hernandez19,Holden15,Cao20} in the temporal dimension, which is known as Temporal Convolutional Networks (TCN)~\cite{Gehring17,Bai18TCN,Luo18}. Here we adopt TCN for future frame prediction due to their performance and robustness, but we encode the space-time body dynamics only with GCN.

\myparagraph{Representation of body joints}
Nearly all literature adopts 3D coordinates or angles. \cite{Julieta17} has noted that encoding residuals of coordinates, thus velocity, may be beneficial. \cite{Mao19LTD,Mao20DCT} has adopted Discrete cosine transform (DCT), thus frequency, which greatly supports for periodic motion. Here we experiment with 3D coordinates and angles, but those representations are compatible with our model.

\myparagraph{Representation of human pose}
Graphs are a natural choice to represent the body. These have mostly been hand-designed, mainly leveraging the natural structure of the kinematic tree~\cite{Jain16,Butepage17,Yan18stgcn}, and encoded via Graph Convolutional Networks (GCN)~\cite{Kipf17gcn}. \cite{Yan18stgcn} learns the adjacency matrix of the graph, still limiting the connectivity to the kinematic tree. Most recently, research has explored all joints linked together and learnt graph edges~\cite{Mao19LTD, Mao20DCT}. Ours also let the training learn a data-driven graph connectivity and edge weights (see Fig.~\ref{fig:spatial_edges} and Sec.~\ref{sec:exp} for an illustration).

\myparagraph{Separable Convolutions}
Separable convolutions~\cite{Szegedy15,Chollet17,Howard17} decouple processing the cross-channel correlations via 1x1 convolutional filters and the spatial correlations via channel-wise spatial convolutions.
These are \emph{depthwise}-separable convolutions, based on the hypothesis that the cross-channel and spatial correlations are sufficiently decoupled, so it is preferable not to map them jointly~\cite{Chollet17}.\\
To the best of our knowledge, only \cite{LaiRejected2018depthwise} and \cite{Balcilar20depthwisespectral} apply this concept to GCNs, but they design different graph edge weights for different channels, in the spatial~\cite{LaiRejected2018depthwise} or spectral domain~\cite{Balcilar20depthwisespectral}. By contrast, our STS-GCN is the first GCN design which separates the graph connectivity itself, by factoring the space-time adjacency matrix. In the spirit of \cite{Chollet17}, our hypothesis is that the space-time cross-talk is limited and that decoupling them is more effective and efficient.

\section{STS-GCN}\label{sec:sts-gcn-main}

The proposed model proceeds by encoding the coordinates of the body joints which are observed in the given input frames and then it leverages the space-time representation to forecast the future joint coordinates. Encoding is modelled by the proposed STS-GCN graph, which considers the interaction of body joints over time, bottleneck'ing the space-time interplay. Decoding future coordinates is modelled with a TCN. In this section, we further provide insights into the STS-GCN model.

\subsection{Problem Formalization}

We observe the body pose of a person, given by the 3D coordinates or angles of its $V$ joints, for $T$ frames. Then we predict the $V$ body joints for the next $K$ future frames.

We denote the joints by 3D vectors $\boldsymbol{x}_{v,k}$ representing joint $v$ at time $k$. The motion history of human poses is denoted by the tensor $\mathcal{X}_{in}=[X_1,X_2...,X_T]$
which we construct out of
matrices of 3D coordinates or angles of joints $X_i \in \mathbb{R}^{3 \times V}$ for frames $i=1\dots T$.
The goal is to predict the future $K$ poses $\mathcal{X}_{out}=[X_{T+1},X_{T+2}...,X_{T+K}]$.

The motion history tensor is encoded into a graph which models the interaction of all body joints across all observed frames. We define the encoding graph $\mathcal{G}=(\mathcal{V},\mathcal{E})$, with $TV$ nodes $i\in\mathcal{V}$, which are all body joints across all observed time frames. Edges
$(i,j)\in\mathcal{E}$ are represented by a spatio-temporal adjacency matrix $A^{st}\in \mathbb{R}^{VT \times VT}$, relating the interactions of all joints at all times.

\subsection{Background on GCN}

The spatio-temporal dependencies of joints across times may be conveniently encoded by a GCN, a graph-based neural network model $f(\mathcal{X}_{in};A,W)$. The input to a graph convolutional layer $l$ is the tensor $\mathcal{H}^{(l)}\in \mathbb{R}^{C^{(l)} \times V \times T}$, which encodes the observed $V$ joints in the $T$ frames. $C^{(l)}$ is the input dimensionality of the hidden representation $\mathcal{H}^{(l)}$. For the first layer, it is $\mathcal{H}^{(1)}=\mathcal{X}_{in}$ and $C^{(1)}=3$.

A graph convolutional layer $l$ outputs the $\mathcal{H}^{(l+1)} \in \mathbb{R}^{C^{(l+1)} \times V \times T}$, given by the following
\begin{equation}
\mathcal{H}^{(l+1)} = \sigma(A^{st-(l)} \mathcal{H}^{(l)} W^{(l)})
\label{eq:gcn}
\end{equation}
where $A^{st-(l)}\in \mathbb{R}^{VT \times VT}$ is the spatio-temporal adjacency matrix of layer $l$, $W^{(l)} \in \mathbb{R}^{C^{(l)} \times C^{(l+1)}}$ are the trainable graph convolutional weights of layer $l$ projecting each graph node from $C^{(l)}$ to $C^{(l+1)}$ dimensions, and $\sigma$ is an activation function such as ReLU, PReLU or tanh.

Two notable graph representations are worth mentioning for their robustness and performance. \cite{Yan18stgcn} constrains the graph encoding to the joint-joint relations, thus to a spatial-only $A^{s}$, only along the kinematic tree, and addresses the time-time relations by a convolutional layer of kernel $T\times T\times 1\times 1$, mapping $T$ frames to $T$ channels.
\cite{Mao20DCT}, the current state-of-the-art in human pose forecasting, also adopts a spatial-only adjacency matrix $A^{s}$, but fully connected.
In both cases, the adjacency matrices are trainable.

\subsection{Space-Time Separable GCN}\label{sec:sts-gcn-sub}

The proposed STS-GCN takes motivation from the interaction of the temporal evolution and the spatial joints, as well as from the belief that the interplay of joint-joint and time-time are privileged. Human pose dynamics depend on 3 types of interactions: \textbf{i.}\ joint-joint; \textbf{ii.}\ time-time; and \textbf{iii.}\ joint-time. STS-GCN allows for all 3 types of interactions, but it bottlenecks the joint-time cross-talk.

The interplay of joints over time is modelled by relating the 3 types of relations within a single spatio-temporal encoding GCN. Bottleneck'ing the space-time cross-talk is realized by factoring the space-time adjacency matrix into the product of separate spatial and temporal adjacency matrices $A^{st}=A^{s}A^{t}$.
A separable space-time graph convolutional layer $l$ is therefore written as follows
\begin{equation}
\mathcal{H}^{(l+1)} = \sigma(A^{s-(l)} A^{t-(l)} \mathcal{H}^{(l)} W^{(l)})
\label{eq:stsgcn}
\end{equation}
where the same notation as in Eq.~\eqref{eq:gcn} applies, apart from the factored $A^{s-(l)} A^{t-(l)}$ of layer $l$ which we explain next.

The adjacency matrix $A^{s}$ is responsible for the joint-joint interplay. It has dimensionality $A^{s}\in \mathbb{R}^{V \times V}$, and it models the full joint-joint relations by trainable $V\times V$ matrices for each instant in time (there are $T$ such matrices).
Similarly $A^{t}$ is responsible for the time-time relations. It has dimensionality $A^{t}\in \mathbb{R}^{T \times T}$ and it defines a full and trainable time-time $T\times T$ relation matrix for each of the $V$ joints.

Note that Eq.~\ref{eq:stsgcn} represents a single GCN layer, encoding the spatio-temporal interplay of the the body dynamics. The factored space-time matrix bottlenecks the space-time cross-talk, it reduces the model parameters and it yields a considerable increase in the forecasting performance, as we illustrate in Sec.~\ref{sec:exp}. Overall, the graph encoding employs four such GCN layers with residual connections PReLU activation functions, cf.~\ref{sec:exp} for the implementation details.

Also note that STS-GCN is the sole human pose forecasting graph encoding which exclusively uses GCNs. This contrasts other competing techniques, mostly encoding time with recurrent neural networks~\cite{Mao20DCT,Mao19LTD,Wang19,Chiu19,Yuan20,Adeli20,Yuan19}, or by the use of convolutional layers with kernels across the temporal dimension~\cite{Yan18stgcn,Cao20}. This is also a key element to parameter efficiency (see Sec.~\ref{sec:exp}.)

\subsection{Discussion on the STS-GCN}

Here we first relate STS-GCN to self-attention mechanisms, then we comment on STS-GCN in relation to most recent work on signed and directed GCNs.

\begin{table*}[]
\centering
\resizebox{.7\textwidth}{!}{
\begin{tabular}{|c|cccc|cccc|cccc|cccc|}
\cline{2-17}
\multicolumn{1}{l|}{} &
  \multicolumn{4}{c|}{Walking} &
  \multicolumn{4}{c|}{Eating} &
  \multicolumn{4}{c|}{Smoking} &
  \multicolumn{4}{c|}{Discussion} \\ \hline
\multicolumn{1}{|c|}{\textit{msec}} &
  {\ul \textit{80}} &
  {\ul \textit{160}} &
  {\ul \textit{320}} &
  {\ul \textit{400}} & 
  {\ul \textit{80}} &
  {\ul \textit{160}} &
  {\ul \textit{320}} &
  {\ul \textit{400}} &
  {\ul \textit{80}} &
  {\ul \textit{160}} &
  {\ul \textit{320}} &
  {\ul \textit{400}} &
  {\ul \textit{80}} &
  {\ul \textit{160}} &
  {\ul \textit{320}} &
  {\ul \textit{400}} \\ \cline{1-17}

ConvSeq2Seq~\cite{Li18} &
  17.7 & 33.5 & 56.3 & 63.6 & 11.0 & 22.4 & 40.7 & 48.4 & 11.6 & 22.8 & 41.3 & 48.9 & 17.1 & 34.5 & 64.8 & 77.6\\
  
LTD-10-10 \cite{Mao19LTD} &
  11.1 &
  21.4 &
  37.3 &
  42.9 &
  7.0 &
  14.8 &
  29.8 &
  37.3 &
  7.5 &
  15.5 &
  30.7 &
  37.5 &
  10.8 &
  24.0 &
  52.7 &
  65.8 \\
DCT-RNN-GCN \cite{Mao20DCT} &
  \textbf{10.0} &
  19.5 &
  34.2 &
  39.8 &
  \textbf{6.4} &
  14.0 &
  28.7 &
  36.2 &
  \textbf{7.0} &
  14.9 &
  29.9 &
  36.4 &
  10.2 &
  23.4 &
  52.1 &
  65.4 \\ \cline{1-17}
Ours &
  10.7 &
  \textbf{16.9} &
  \textbf{29.1} &
  \textbf{32.9} &
  6.8 &
  \textbf{11.3} &
  \textbf{22.6} &
  \textbf{25.4} &
  7.2 &
  \textbf{11.6} &
  \textbf{22.3} &
  \textbf{25.8} &
  \textbf{9.8} &
  \textbf{16.8} &
  \textbf{33.4} &
  \textbf{40.2} \\ \bottomrule
\end{tabular}}
\end{table*}

\begin{table*}[]
\centering
\resizebox{\textwidth}{!}{
\begin{tabular}{|c|cccc|cccc|cccc|cccc|cccc|cccc|}
\cline{2-25}
\multicolumn{1}{l|}{} &
  \multicolumn{4}{c|}{Directions} &
  \multicolumn{4}{c|}{Greeting} &
  \multicolumn{4}{c|}{Phoning} &
  \multicolumn{4}{c|}{Posing} &
  \multicolumn{4}{c|}{Purchases} &
  \multicolumn{4}{c|}{Sitting} \\ \hline
\multicolumn{1}{|c|}{\textit{msec}} &
  {\ul \textit{80}} &
  {\ul \textit{160}} &
  {\ul \textit{320}} &
  {\ul \textit{400}} &
  {\ul \textit{80}} &
  {\ul \textit{160}} &
  {\ul \textit{320}} &
  {\ul \textit{400}} &
  {\ul \textit{80}} &
  {\ul \textit{160}} &
  {\ul \textit{320}} &
  {\ul \textit{400}} &
  {\ul \textit{80}} &
  {\ul \textit{160}} &
  {\ul \textit{320}} &
  {\ul \textit{400}} &
  {\ul \textit{80}} &
  {\ul \textit{160}} &
  {\ul \textit{320}} &
  {\ul \textit{400}} &
  {\ul \textit{80}} &
  {\ul \textit{160}} &
  {\ul \textit{320}} &
  {\ul \textit{400}} \\ \cline{1-25}
ConvSeq2Seq \cite{Li18} &
 13.5 & 29.0 & 57.6 & 69.7 & 22.0 & 45.0 & 82.0 & 96.0 & 13.5 & 26.6 & 49.9 & 59.9 & 16.9 & 36.7 & 75.7 & 92.9 & 20.3 & 41.8 & 76.5 & 89.9 & 13.5 & 27.0 & 52.0 & 63.1 \\

LTD-10-10 \cite{Mao19LTD} &
  8.0 &
  18.8 &
  43.7 &
  54.9 &
  14.8 &
  31.4 &
  65.3 &
  79.7 &
  9.3 &
  19.1 &
  39.8 &
  49.7 &
  10.9 &
  25.1 &
  59.1 &
  75.9 &
  13.9 &
  30.3 &
  62.2 &
  75.9 &
  9.8 &
  20.5 &
  44.2 &
  55.9 \\
DCT-RNN-GCN \cite{Mao20DCT} &
  \textbf{7.4} &
  18.5 &
  44.5 &
  56.5 &
  13.7 &
  30.1 &
  63.8 &
  78.1 &
  8.6 &
  18.3 &
  39.0 &
  49.2 &
  10.2 &
  24.2 &
  58.5 &
  75.8 &
  13.0 &
  29.2 &
  60.4 &
  73.9 &
  9.3 &
  20.1 &
  44.3 &
  56.0 \\ \cline{1-25}
\multicolumn{1}{|c|}{Ours} &
  \textbf{7.4} &
  \textbf{13.5} &
  \textbf{29.2} &
  \textbf{34.7} &
  \textbf{12.4} &
  \textbf{21.8} &
  \textbf{42.1} &
  \textbf{49.2} &
  \textbf{8.2} &
  \textbf{13.7} &
  \textbf{26.9} &
  \textbf{30.9} &
  \textbf{9.9} &
  \textbf{18.0} &
  \textbf{38.2} &
  \textbf{45.6} &
  \textbf{11.9} &
  \textbf{21.3} &
  \textbf{42.0} &
  \textbf{48.7} &
  \textbf{9.1} &
  \textbf{15.1} &
  \textbf{29.9} &
  \textbf{35.0} \\ \bottomrule
\end{tabular}}
\end{table*}

\begin{table*}[h!]
\centering
\resizebox{\textwidth}{!}{
\begin{tabular}{|c|cccc|cccc|cccc|cccc|cccc|cccc|}
\cline{2-25}
\multicolumn{1}{l|}{} &
  \multicolumn{4}{c|}{Sitting Down} &
  \multicolumn{4}{c|}{Taking Photo} &
  \multicolumn{4}{c|}{Waiting} &
  \multicolumn{4}{c|}{Walking Dog} &
  \multicolumn{4}{c|}{Walking Together} &
  \multicolumn{4}{c|}{Average} \\ \hline
\multicolumn{1}{|c|}{\textit{msec}} &
  {\ul \textit{80}} &
  {\ul \textit{160}} &
  {\ul \textit{320}} &
  {\ul \textit{400}} &
  {\ul \textit{80}} &
  {\ul \textit{160}} &
  {\ul \textit{320}} &
  {\ul \textit{400}} &
  {\ul \textit{80}} &
  {\ul \textit{160}} &
  {\ul \textit{320}} &
  {\ul \textit{400}} &
  {\ul \textit{80}} &
  {\ul \textit{160}} &
  {\ul \textit{320}} &
  {\ul \textit{400}} &
  {\ul \textit{80}} &
  {\ul \textit{160}} &
  {\ul \textit{320}} &
  {\ul \textit{400}} &
  {\ul \textit{80}} &
  {\ul \textit{160}} &
  {\ul \textit{320}} &
  {\ul \textit{400}} \\ \cline{1-25}

ConvSeq2Seq \cite{Li18} &
 20.7 & 40.6 & 70.4 & 82.7 & 12.7 & 26.0 & 52.1 & 63.6 & 14.6 & 29.7 & 58.1 & 69.7 & 27.7 & 53.6 & 90.7 & 103.3 & 15.3 & 30.4 & 53.1 & 61.2 & 16.6 & 33.3 & 61.4 & 72.7\\

LTD-10-10 \cite{Mao19LTD} &
  15.6 &
  31.4 &
  59.1 &
  71.7 &
  8.9 &
  18.9 &
  41.0 &
  51.7 &
  9.2 &
  19.5 &
  43.3 &
  54.4 &
  20.9 &
  40.7 &
  73.6 &
  86.6 &
  9.6 &
  19.4 &
  36.5 &
  44.0 &
  11.2 &
  23.4 &
  47.9 &
  58.9 \\
DCT-RNN-GCN \cite{Mao20DCT} &
  14.9 &
  30.7 &
  59.1 &
  72.0 &
  8.3 &
  18.4 &
  40.7 &
  51.5 &
  8.7 &
  19.2 &
  43.4 &
  54.9 &
  20.1 &
  40.3 &
  73.3 &
  86.3 &
  8.9 &
  18.4 &
  35.1 &
  41.9 &
  10.4 &
  22.6 &
  47.1 &
  58.3 \\ \cline{1-25}
\multicolumn{1}{|c|}{Ours} &
  \textbf{14.4} &
  \textbf{23.7} &
  \textbf{41.9} &
  \textbf{47.9} &
  \textbf{8.2} &
  \textbf{14.2} &
  \textbf{29.7} &
  \textbf{33.6} &
  \textbf{8.6} &
  \textbf{14.7} &
  \textbf{29.6} &
  \textbf{35.2} &
  \textbf{17.6} &
  \textbf{29.4} &
  \textbf{52.6} &
  \textbf{59.6} &
  \textbf{8.6} &
  \textbf{14.3} &
  \textbf{26.5} &
  \textbf{30.5} &
  \textbf{10.1} &
  \textbf{17.1} &
  \textbf{33.1} &
  \textbf{38.3} \\ \bottomrule
\end{tabular}}
\caption{MPJPE error in mm for short-term prediction of 3D joint positions on Human3.6M. Our model outperforms the state-of-the-art by a large margin. The margin is smaller for very-short-term predictions on periodic actions, e.g.\ 2-4 frame (80-160 msec) for \emph{Walking} and \emph{Eating}. 
The margin is larger for the more challenging case of longer-term and aperiodic actions, e.g. up to 40\% for \emph{Posing} at 10-frame (400 msec). 
See Sec.~\ref{sec:exp_sota} for the discussion.
}
\label{tab:short_mm}
\end{table*}

\myparagraph{Separable graph convolutions and self-attention}
Most recent pose forecasting work has leveraged self-attention to encode the relation of frames~\cite{Mao20DCT,Cai20} and/or the relation of joints~\cite{Cai20}. Here we relate the proposed STS-GCN to the self-attention mechanisms of~\cite{Mao20DCT,Cai20}. Finally we relate these to Graph Attention Networks (GAT)~\cite{Velickovic18gat}.

Let us first re-write part of the GCN layer of Eq.~\ref{eq:gcn} with the Einstein summation, omitting the indication of layer $l$, the projection matrix $W$ and the non-linearity $\sigma$ for better clarity of notation:
\begin{equation}\label{eq:full_stsgcn_einsum}
A^{st} \mathcal{H} = \sum_{vm}  A^{st}_{wkvm} \mathcal{H}_{vmc}
\end{equation}
having explicitly indicated with indexes the dimensions of $A^{st} \in \mathbb{R}^{VT \times VT}$ and $\mathcal{H}\in \mathbb{R}^{C \times V \times T}$, i.e.\ indexing spatial joints as $v,w=1,...,V$ and times with $m,k=1,...,T$.

Let us now re-write the corresponding part of the STS-GCN layer of Eq.~\ref{eq:stsgcn} with the Einstein summation, again omitting the projection matrix $W$ and the non-linearity $\sigma$ for clarity of notation:
\begin{equation}\label{eq:stsgcn_einsum}
A^{s} (A^{t} \mathcal{H}) = \sum_{v} A^{s}_{wkv} (\sum_{m} A^{t}_{kvm} \mathcal{H}_{vmc})
= \sum_{v} A^{s}_{wkv} \mathcal{H}^{t}_{kvc}
\end{equation}

where, as above, we have indicated indexes for $A^{s}\in \mathbb{R}^{V \times V}$ (for each of the T times) and $A^{t}\in \mathbb{R}^{T \times T}$ (for each of the V joints) as $v,w=1,...,V$ for the spatial joints and as $m,k=1,...,T$ for the times.

Let us now turn to the current best technique for pose forecasting~\cite{Mao20DCT}. They adopt a GCN for modelling the spatial interaction of joints at the same time, which coincides with the rightmost term in Eq.~\ref{eq:stsgcn_einsum}.

Their temporal modelling is however different from ours, as they adopt an attention formulation $\sigma(QK)V$. Writing it with the Einstein summation yields:
\begin{equation}\label{eq:time_atten_einsum}
\sum_m \left(\sum_{c} Q^{t}_{kvc} {K^{t}_{vcm}} \right) V_{vmc} = \sum_m A^{QK-t}_{kvm} V_{vmc}
\end{equation}

Comparing the right term of Eq.~\ref{eq:time_atten_einsum} with the separable temporal GCN (the term within parentheses in Eq.~\ref{eq:stsgcn_einsum}), we note that the approach of \cite{Mao20DCT}, modelling space and time with the different mechanisms of GCN and attention, may also be explained as a separable space-time GCN. The main difference is that $A^{QK-t}$ is a function of the product of inner representation vectors, both stemming from $\mathcal{H}$. By contrast our temporal adjacency matrix $A^{t}$ learns the specific pair-wise interaction of relative time shifts. Similar arguments apply when comparing the proposed STS-GCN with the recent GAT~\cite{Velickovic18gat}. We evaluate the difference \emph{wrt}~\cite{Mao20DCT} quantitatively and conduct ablation studies on the adjacency matrices in Sec.~\ref{sec:exp}.

\myparagraph{Signed and Directed GCNs}
Let us now consider that the adjacency matrix $A^{st}$ and its factored terms, $A^{s}$ and $A^{t}$, are trainable parameters. Adjacency matrices were similarly trained by \cite{Mao20DCT}, which considers a fully connected matrix, and by \cite{Yan18stgcn}, which defines specific learnable parameters (denoted $M$ in \cite{Yan18stgcn}) to multiply the manually-constructed graph (based on the kinematic tree and the sequential time connections). When encoding the spatio-temporal body dynamics, trainable parameters yield better performance and match the intuition, i.e.\ they learn the interaction between specific joints and at certain relative temporal offsets.

Trainable parameters result in signed and directed GCNs (see Figs.~\ref{fig:spatial_edges} for an illustration). Both aspects have been surveyed recently~\cite{Bronstein17,Wu20}. In particular, recent work from~\cite{Tong20,Li2018diff,Atwood16} maintain that directed graphs encode richer information from their neighborhood, instead of being limited to distance ranges. Similarly, recent work from~\cite{Derr18} demonstrate the superior performance of signed GCNs.

Following the classification of~\cite{Bronstein17,Wu20}, the proposed STS-GCN and the GCNs of \cite{Mao20DCT,Yan18stgcn} are spatial GCNs. This follows from their non-symmetric and possibly ill-posed signed Laplacian matrices, which do not have orthogonal eigendecompositions and are not easily interpretable by spectral-domain constructions~\cite{Bronstein17}. We maintain this makes an interesting direction for future investigation, only partly addressed by very recent work~\cite{Veerman20}.

\subsection{Decoding future coordinates}

Given the encoded observed body dynamics, the estimation of the 3D coordinates or angles of the body joints in the future is delegated to convolutional layers applying to the temporal dimension. These map the observed frames into the future horizon and refine the estimates via a multi-layered architecture.

Altogether, these layers make a decoder which is generally dubbed Temporal Convolutional Networks (TCN)~\cite{Gehring17,Bai18TCN,Luo18}. While several other sequence modelling options are available, including LSTM~\cite{lstm97}, GRU~\cite{gru14} and Transformer Networks~\cite{giuliari20}, here we adopt TCNs for their simplicity and robustness, further to satisfactory performance~\cite{Li18}.


\begin{table*}[]
\centering
\resizebox{0.7\textwidth}{!}{
\begin{tabular}{|c|cccc|cccc|cccc|cccc|}
\cline{2-17}
\multicolumn{1}{l|}{\textbf{}} &
  \multicolumn{4}{c|}{Walking} &
  \multicolumn{4}{c|}{Eating} &
  \multicolumn{4}{c|}{Smoking} &
  \multicolumn{4}{c|}{Discussion} \\ \hline
\multicolumn{1}{|c|}{\textit{msec}} &
  {\ul \textit{560}} &
  {\ul \textit{720}} &
  {\ul \textit{880}} &
  {\ul \textit{1000}} &
  {\ul \textit{560}} &
  {\ul \textit{720}} &
  {\ul \textit{880}} &
  {\ul \textit{1000}} &
  {\ul \textit{560}} &
  {\ul \textit{720}} &
  {\ul \textit{880}} &
  {\ul \textit{1000}} &
  {\ul \textit{560}} &
  {\ul \textit{720}} &
  {\ul \textit{880}} &
  {\ul \textit{1000}} \\ \cline{1-17}
ConvSeq2Seq \cite{Li18}& 
 72.2 &  77.2 &  80.9 &  82.3 &  61.3  & 72.8  & 81.8  & 87.1 &  60.0 &  69.4 &  77.2 &  81.7 &  98.1  & 112.9  & 123.0 &  129.3\\

LTD-50-25 \cite{Mao19LTD}&
 50.7 & 54.4 & 57.4 & 60.3 & 51.5 & 62.6 & 71.3 & 75.8 & 50.5 & 59.3 & 67.1 & 72.1 & 88.9 & 103.9 & 113.6 & 118.5\\

LTD-10-25 \cite{Mao19LTD}& 51.8 & 56.2 & 58.9 & 60.9 & 50.0 & 61.1 & 69.6 & 74.1 & 51.3 & 60.8 & 68.7 & 73.6 & 87.6 & 103.2 & 113.1 & 118.6 \\
LTD-10-10 \cite{Mao19LTD}& 53.1 & 59.9 & 66.2 & 70.7 & 51.1 & 62.5 & 72.9 & 78.6 & 49.4 & 59.2 & 66.9 & 71.8 & 88.1 & 104.4 & 115.5 & 121.6 \\
DCT-RNN-GCN \cite{Mao20DCT} & 47.4 & 52.1 & 55.5 & 58.1 & 50.0 & 61.4 & 70.6 & 75.5 & 47.5 & 56.6 & 64.4 & 69.5 & 86.6 & 102.2 & 113.2 & 119.8 \\ \cline{1-17}
\multicolumn{1}{|c|}{Ours} &
  \textbf{40.6} &
  \textbf{45.0} &
  \textbf{48.0} &
  \textbf{51.8} &
  \textbf{33.9} &
  \textbf{40.2} &
  \textbf{46.2} &
  \textbf{52.4} &
  \textbf{33.6} &
  \textbf{39.6} &
  \textbf{45.4} &
  \textbf{50.0} &
  \textbf{53.4} &
  \textbf{63.6} &
  \textbf{72.3} &
  \textbf{78.8} \\ \bottomrule
\end{tabular}}
\end{table*}

\begin{table*}[]
\centering
\resizebox{\textwidth}{!}{
\begin{tabular}{|c|cccc|cccc|cccc|cccc|cccc|cccc|}
\cline{2-25}
\multicolumn{1}{l|}{} &
  \multicolumn{4}{c|}{Directions} &
  \multicolumn{4}{c|}{Greeting} &
  \multicolumn{4}{c|}{Phoning} &
  \multicolumn{4}{c|}{Posing} &
  \multicolumn{4}{c|}{Purchases} &
  \multicolumn{4}{c|}{Sitting} \\ \hline
\multicolumn{1}{|c|}{\textit{msec}} &
  {\ul \textit{560}} &
  {\ul \textit{720}} &
  {\ul \textit{880}} &
  {\ul \textit{1000}} &
  {\ul \textit{560}} &
  {\ul \textit{720}} &
  {\ul \textit{880}} &
  {\ul \textit{1000}} &
  {\ul \textit{560}} &
  {\ul \textit{720}} &
  {\ul \textit{880}} &
  {\ul \textit{1000}} &
  {\ul \textit{560}} &
  {\ul \textit{720}} &
  {\ul \textit{880}} &
  {\ul \textit{1000}} &
  {\ul \textit{560}} &
  {\ul \textit{720}} &
  {\ul \textit{880}} &
  {\ul \textit{1000}} &
  {\ul \textit{560}} &
  {\ul \textit{720}} &
  {\ul \textit{880}} &
  {\ul \textit{1000}} \\ \cline{1-25}
ConvSeq2Seq \cite{Li18} &
 86.6 & 99.8 & 109.9 & 115.8 & 116.9 & 130.7 & 142.7 & 147.3 & 77.1 & 92.1 & 105.5 & 114.0 & 122.5 & 148.8 & 171.8 & 187.4 & 111.3 & 129.1 & 143.1 & 151.5 & 82.4 & 98.8 & 112.4 & 120.7\\
 
LTD-50-25 \cite{Mao19LTD} &  
 74.2 & 88.1 & 99.4 & 105.5 & 104.8 & 119.7 & 132.1 & 136.8 & 68.8 & 83.6 & 96.8 & 105.1 & 110.2 & 137.8 & 160.8 & 174.8 & 99.2 & 114.9 & 127.1 & 134.9 & 79.2 & 96.2 & 110.3 & 118.7\\

LTD-10-25 \cite{Mao19LTD} &
  76.1 &
  91.0 &
  102.8 &
  108.8 &
  104.3 &
  120.9 &
  134.6 &
  140.2 &
  68.7 &
  84.0 &
  97.2 &
  105.1 &
  109.9 &
  136.8 &
  158.3 &
  171.7 &
  99.4 &
  114.9 &
  127.9 &
  135.9 &
  78.5 &
  95.7 &
  110.0 &
  118.8 \\
LTD-10-10 \cite{Mao19LTD} &
  72.2 &
  86.7 &
  98.5 &
  105.8 &
  103.7 &
  120.6 &
  134.7 &
  140.9 &
  67.8 &
  83.0 &
  96.4 &
  105.1 &
  107.6 &
  136.1 &
  159.5 &
  175.0 &
  98.3 &
  115.1 &
  130.1 &
  139.3 &
  76.4 &
  93.1 &
  106.9 &
  115.7 \\
DCT-RNN-GCN \cite{Mao20DCT} &
  73.9 &
  88.2 &
  100.1 &
  106.5 &
  101.9 &
  118.4 &
  132.7 &
  138.8 &
  67.4 &
  82.9 &
  96.5 &
  105.0 &
  107.6 &
  136.8 &
  161.4 &
  178.2 &
  95.6 &
  110.9 &
  125.0 &
  134.2 &
  76.4 &
  93.1 &
  107.0 &
  115.9 \\ \cline{1-25}
\multicolumn{1}{|c|}{Ours} &
  \textbf{47.6} &
  \textbf{56.5} &
  \textbf{64.5} &
  \textbf{71.0} &
  \textbf{64.8} &
  \textbf{76.3} &
  \textbf{85.5} &
  \textbf{91.6} &
  \textbf{41.8} &
  \textbf{51.1} &
  \textbf{59.3} &
  \textbf{66.1} &
  \textbf{64.3} &
  \textbf{79.3} &
  \textbf{94.5} &
  \textbf{106.4} &
  \textbf{63.7} &
  \textbf{74.9} &
  \textbf{86.2} &
  \textbf{93.5} &
  \textbf{47.7} &
  \textbf{57.0} &
  \textbf{67.4} &
  \textbf{75.2} \\ \bottomrule
\end{tabular}}
\end{table*}

\begin{table*}[h!]
\centering
\resizebox{\textwidth}{!}{
\begin{tabular}{|c|cccc|cccc|cccc|cccc|cccc|cccc|}
\cline{2-25}
\multicolumn{1}{l|}{} &
  \multicolumn{4}{c|}{Sitting Down} &
  \multicolumn{4}{c|}{Taking Photo} &
  \multicolumn{4}{c|}{Waiting} &
  \multicolumn{4}{c|}{Walking Dog} &
  \multicolumn{4}{c|}{Walking Together} &
  \multicolumn{4}{c|}{Average} \\ \hline
\multicolumn{1}{|c|}{\textit{msec}} &
  {\ul \textit{560}} &
  {\ul \textit{720}} &
  {\ul \textit{880}} &
  {\ul \textit{1000}} &
  {\ul \textit{560}} &
  {\ul \textit{720}} &
  {\ul \textit{880}} &
  {\ul \textit{1000}} &
  {\ul \textit{560}} &
  {\ul \textit{720}} &
  {\ul \textit{880}} &
  {\ul \textit{1000}} &
  {\ul \textit{560}} &
  {\ul \textit{720}} &
  {\ul \textit{880}} &
  {\ul \textit{1000}} &
  {\ul \textit{560}} &
  {\ul \textit{720}} &
  {\ul \textit{880}} &
  {\ul \textit{1000}} &
  {\ul \textit{560}} &
  {\ul \textit{720}} &
  {\ul \textit{880}} &
  {\ul \textit{1000}} \\ \cline{1-25}
ConvSeq2Seq \cite{Li18} &
106.5 & 125.1 & 139.8 & 150.3 & 84.4 & 102.4 & 117.7 & 128.1 & 87.3 & 100.3 & 110.7 & 117.7 & 122.4 & 133.8 & 151.1 & 162.4 & 72.0 & 77.7 & 82.9 & 87.4 & 90.7 & 104.7 & 116.7 & 124.2\\
 
LTD-50-25 \cite{Mao19LTD} &
 100.2 & 118.2 & 133.1 & 143.8 & 75.3 & 93.5 & 108.4 & 118.8 & 77.2 & 90.6 & 101.1 & 108.3 & 107.8 & 120.3 & 136.3 & 146.4 & 56.0 & 60.3 & 63.1 & 65.7 & 79.6 & 93.6  & 105.2 & 112.4\\

LTD-10-25 \cite{Mao19LTD} &
  99.5 & 118.5 & 133.6 & 144.1 & 76.8 & 95.3 & 110.3 & 120.2 & 75.1 & 88.7 & 99.5 & 106.9 & 105.8 & 118.7 & 132.8 & 142.2 & 58.0 & 63.6 & 67.0 & 69.6 & 79.5 & 94.0 & 105.6 & 112.7\\
LTD-10-10  \cite{Mao19LTD}&
  96.2 &
  115.2 &
  130.8 &
  142.2 &
  72.5 &
  90.9 &
  105.9 &
  116.3 &
  73.4 &
  88.2 &
  99.8 &
  107.5 &
  109.7 &
  122.8 &
  139.0 &
  150.1 &
  55.7 &
  61.3 &
  66.4 &
  69.8 &
  78.3 &
  93.3 &
  106.0 &
  114.0 \\
DCT-RNN-GCN \cite{Mao20DCT} &
  97.0 &
  116.1 &
  132.1 &
  143.6 &
  72.1 &
  90.1 &
  105.5 &
  115.9 &
  74.5 &
  89.0 &
  100.3 &
  108.2 &
  108.2 &
  120.6 &
  135.9 &
  146.9 &
  52.7 &
  57.8 &
  62.0 &
  64.9 &
  77.3 &
  91.8 &
  104.1 &
  112.1 \\ \cline{1-25}
\multicolumn{1}{|c|}{Ours} &
  \textbf{63.3} &
  \textbf{73.9} &
  \textbf{86.2} &
  \textbf{94.3} &
  \textbf{47.0} &
  \textbf{57.4} &
  \textbf{67.2} &
  \textbf{76.9} &
  \textbf{47.3} &
  \textbf{56.8} &
  \textbf{66.1} &
  \textbf{72.0} &
  \textbf{74.7} &
  \textbf{85.7} &
  \textbf{96.2} &
  \textbf{102.6} &
  \textbf{38.9} &
  \textbf{44.0} &
  \textbf{48.2} &
  \textbf{51.1} &
  \textbf{50.8} &
  \textbf{60.1} &
  \textbf{68.9} &
  \textbf{75.6} \\ \bottomrule
\end{tabular}}
\caption{MPJPE error in mm  for long-term prediction of 3D joint positions on Human3.6M. Our model outperforms the state-of-the-art by a large margin for each time prediction horizon and each action. Largest improvements \emph{wrt} the current best~\cite{Mao20DCT} are obtained for the most challenging cases of longer-term predictions (22-25 frame, 880-1000 msec) of aperiodic actions such as \emph{Sitting} (36\%), \emph{Phoning} (43\%) and \emph{Posing} (40\%).
The average improvement over the 14-25 frame (560-1000 msec) predictions is 34\%. See Sec.~\ref{sec:exp_sota} for the discussion.
%
%
}
\label{tab:long_mm}
\end{table*}

\subsection{Training}\label{sec:model-train}

The proposed architecture is trained end-to-end supervisedly. Supervision is provided by either of the losses that measure error \emph{wrt} ground truth in terms of Mean Per Joint Position Error (MPJPE)~\cite{Human36M,Mao19LTD} and Mean Angle Error (MAE)~\cite{Julieta17,Li18,Gui18,Wang19,Mao20DCT}.
The loss based on MPJPE is:
\begin{equation}
L_{MPJPE}= \frac{1}{V(T+K)} \sum_{k=1}^{T+K} \sum_{v=1}^{V} ||\hat{\boldsymbol{x}}_{vk}-\boldsymbol{x}_{vk} ||_{2}
\end{equation}
where $\hat{\boldsymbol{x}}_{vk}  \in \mathbb{R}^{3} $ denotes the predicted coordinates of the joint $v$ in the frame $k$ and $\boldsymbol{x}_{vk} \in \mathbb{R}^{3}$ is the corresponding ground truth.
The loss based on MAE is given by:
\begin{equation}
L_{MAE}= \frac{1}{V(T+K)} \sum_{k=1}^{T+K} \sum_{v=1}^{V} |\hat{\boldsymbol{x}}_{vk}-\boldsymbol{x}_{vk} |
\label{eq:mae_loss}
\end{equation}
where $\hat{\boldsymbol{x}}_{vk} \in \mathbb{R}^{3} $ denotes the predicted joint angles in exponential map representation of the joint $v$ in the frame $k$ and $\boldsymbol{x}_{vk} \in \mathbb{R}^{3} $ is its ground truth.

\section{Experimental evaluation}\label{sec:exp}

We experimentally evaluate the proposed model against the state-of-the-art on three recent, large-scale and challenging benchmarks, Human3.6M~\cite{Human36M}, AMASS~\cite{AMASS} and 3DPW~\cite{3DPW}. Additionally we conduct ablation studies, evaluate the model qualitatively and illustrate what spatio-temporal graph $\mathcal{G}$ is trained from data.

\subsection{Datasets and metrics}

\myparagraph{Human3.6M~\cite{Human36M}} The dataset is wide-spread for human pose forecasting and large, consisting of 3.6 million 3D human poses and the corresponding images. It consists of 7 actors performing 15 different actions (e.g.\ \emph{Walking}, \emph{Eating}, \emph{Phoning}).  The actors are represented as skeletons of 32 joints. The orientation of joints are represented as exponential maps, from which the 3D coordinates may be computed~\cite{Taylor07, Fragkiadaki15}. For each pose, we consider 22 joints out of the provided 32 for estimating MPJPE and 16 for the MAE. Following the current literature \cite{Mao19LTD, Mao20DCT, Julieta17}, we use the subject 11 (S11) for validation, the subject 5 (S5) for testing, and all the rest of the subjects for training.

\myparagraph{AMASS~\cite{AMASS}} The Archive of Motion Capture as Surface Shapes (AMASS) dataset has been recently proposed, to gather 18 existing mocap datasets.  Following \cite{Mao20DCT}, we select 13 from those and take 8 for training, 4 for validation and 1 (\textit{BMLrub}) as the test set. Then we use the SMPL~\cite{Loper15} parameterization to derive a representation of human pose based on a shape vector, which defines the human skeleton, and its joints rotation angles. 
We obtain human poses in 3D by applying forward kinematics. Overall, AMASS consists of 40 human-subjects that perform the action of walking. Each human pose is represented by 52 joints, including 22 body joints and 30 hand joints. Here we consider for forecasting the body joints only and discard from those 4 static ones, leading to an 18-joint human pose. As for \cite{Human36M}, also these sequences are downsampled to 25 fps.

\myparagraph{3DPW~\cite{3DPW}} The 3D Pose in the Wild dataset \cite{3DPW} consists of video sequences acquired by a moving phone camera. 3DPW includes indoor and outdoor actions. Overall, it contains 51,000 frames captured at 30Hz, divided into 60 video sequences. We use this dataset to test generalization of the models which we train AMASS.

\myparagraph{Metrics}
Following the benchmark protocols, we adopt the MPJPE and MAE error metrics (see Sec.~\ref{sec:model-train}). The first quantifies the error of the 3D coordinate predictions in mm. The second measures the angle error in degrees. We follow the protocol of \cite{Mao19LTD} and compute MAE with Euler angles. Due to this representation, MAE suffers from an inherent ambiguity, and MPJPE is more effective~\cite{Cai20, Akhter15}, so mostly adopted here.

\myparagraph{Implementation details} 
The graph encoding is given by 4 layers of STS-GCN, which only differ in the number of channels $C^{(l)}$: from 3 (the input 3D coordinates x,y,z or angles), to 64, then 32, 64 and finally 3 (cf.\ Sec.~\ref{sec:sts-gcn-sub}), by means of the projection matrices $W^{(l)}$.
At each layer we adopt batch normalization \cite{batchnorm15} and residual connections.
Our code is in Pytorch and uses ADAM~\cite{adam15} as optimizer. The learning rate is set to 0.01 and decayed by a factor of 0.1 every 5 epochs after the $20{\text{-th}}$. The batch size is 256. On Human3.6M, training for 30 epochs on an NVIDIA RTX 2060 GPU takes 20 minutes.

\subsection{Comparison to the state-of-the-art}\label{sec:exp_sota}

\begin{figure*}
\centering
{\includegraphics[width=1\textwidth]{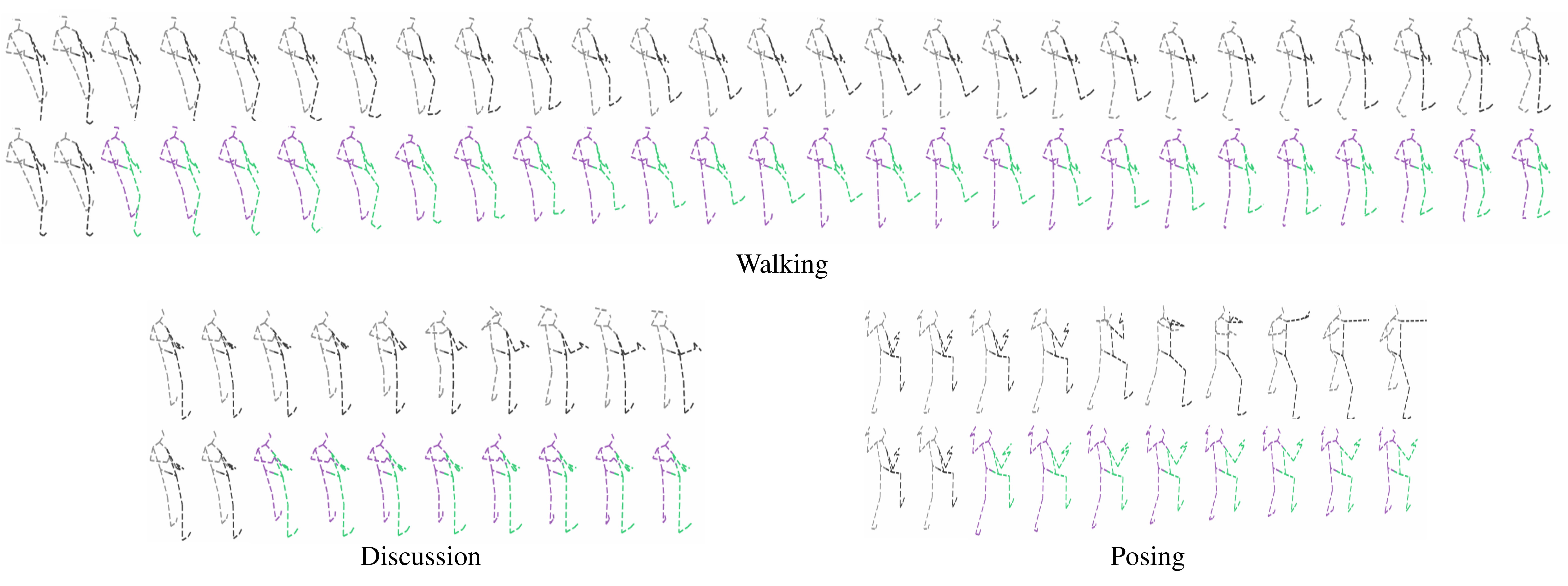} }
\caption{
Sample long-term predictions (25 frames, 1 sec) for the actions of \emph{Walking}, \emph{Discussion} and \emph{Posing}.
(One every three frames are shown for the second two.)
Purple/green limbs are the left/right sides of the body. Gray/black pictorials indicate the observed ground-truth (GT) skeletons. Predictions accurately match the GT. Mistakes may be observed on the left hand of the person in \emph{Discussion} and in the aperiodic motion of \emph{Posing}, an action performed in different ways in the training dataset. Zoom in for details.
}\label{fig:seq_pred}
\end{figure*}

We quantitatively evaluate our proposed model against the state-of-the-art both for short-term ($<$500 msec) and long-term ($>$500 msec) predictions.

We include into the comparison: ConvSeq2Seq~\cite{Li18}, which adopts convolutional layers, separately encoding long- and short-term history;
LTD-X-Y~\cite{Mao19LTD}, which encodes the sequence frequency with a DCT, prior to a GCN (X and Y stand for the number of observed and predicted frames);
BC-WGAIL-div~\cite{Wang19}, adopting reinforcement learning;
and finally DCT-RNN-GCN~\cite{Mao20DCT}, the current best performer, which extends LTD-X-Y with an RNN and motion-attention.

All algorithms take as input 10 frames (400 msec), with the exception of LTD, for which we also report the case of larger number of input frames. Then algorithms predict future poses for the next 2 to 10 frames (80-400 msec) in the case of short-term, and for 14-25 frames (560-1000 msec) in the long-term case.

\myparagraph{Human3.6M: 3D Joint Positions}
Let us consider Tables \ref{tab:short_mm} and \ref{tab:long_mm} for the tests on short- and long-term prediction respectively. Across all time horizons in both tables, our model outperforms all competing techniques, with the only exception of 3 experiments out of 120 (2-frame predictions for \emph{Walking}, \emph{Eating} and \emph{Directions}), where it is within a marginal error. 

Considering the average errors in Table~\ref{tab:short_mm}, the improvement of our model over the current best~\cite{Mao20DCT} ranges from 3\% in the case of 2 time frames, up to 34\% improvement for the more challenging case of 10 frames. 
Note that, at the 10-frame horizon, improvements are less in the case of periodic actions such as \emph{Walking} (17\%) but larger for aperiodic actions such as \emph{Posing} (40\%). We believe this is because of the DCT encoding of \cite{Mao20DCT}.

We illustrate in Table~\ref{tab:long_mm} the more ardous long-term prediction horizons. Our predictions at 560 msec (14 frames) are more accurate than those of \cite{Mao20DCT} by 27 mm, while at 1 sec (25 frames) our model reaches an improvement of 37 mm. In average across predictions over 14-25 frames, our model outperforms the current best \cite{Mao20DCT} by 34\%.

\begin{table}[h!]
\centering
\resizebox{0.49\textwidth}{!}{
\begin{tabular}{|c|cccccccc|}
\cline{2-9}

\multicolumn{1}{l|}{} &
  \multicolumn{8}{c|}{Average} \\ \hline

\multicolumn{1}{|c|}{\textit{msec}} &
  {\ul \textit{80}} &
  {\ul \textit{160}} &
  {\ul \textit{320}} &
  {\ul \textit{400}} &
  {\ul \textit{560}} &
  {\ul \textit{720}} &
  {\ul \textit{880}} &
  {\ul \textit{1000}} \\ \cline{1-9}
LTD-10-25 \cite{Mao19LTD} &
  0.34 & 0.57 & 0.93 & 1.06 &  1.27 & 1.44 & 1.57 & 1.66\\

LTD-10-10 \cite{Mao19LTD} &
  0.32 & 0.55 & 0.91 & 1.04 & 1.26 & 1.44 & 1.59 & 1.68  
 \\

BC-WGAIL-div \cite{Wang19} &
  0.31 &
  0.57 &
  0.90 &
  1.02 &
  1.23 &
  - &
  - &
  1.65 \\

DCT-RNN-GCN \cite{Mao20DCT} &
  0.31 & 0.55 & 0.90 & 1.04 & 1.25 & 1.42  & 1.56 & 1.65\\ \cline{1-9}

\multicolumn{1}{|c|}{Ours} &
  \textbf{0.24} &
  \textbf{0.39} &
  \textbf{0.59} &
  \textbf{0.66} &
  \textbf{0.79} &
  \textbf{0.92} &
  \textbf{1.00} &
  \textbf{1.09} \\ \bottomrule
\end{tabular}}
\caption{Average MAE prediction errors over all actions of Human3.6M. Our model improves consistently over the state-of-the-art by a large margin. 
%
}
\label{tab:angle_avg}
\end{table}

\myparagraph{Human3.6M: Joint Angles} Average angle errors are reported in Table \ref{tab:angle_avg}. Our model outperforms the current best \cite{Mao20DCT} with larger improvements on the long-term horizon. The performance increase is 23\% for 2 frames and it is 34\% for 25 future frames.

\begin{table}[h!]
\centering
\resizebox{0.49\textwidth}{!}{
\begin{tabular}{|c|cccccccc|}
\cline{2-9}
\multicolumn{1}{c|}{} & \multicolumn{8}{c|}{AMASS-BMLrub}                     \\ \hline
\multicolumn{1}{|c|}{\textit{msec}} &
  {\ul \textit{80}} &
  {\ul \textit{160}} &
  {\ul \textit{240}} &
  {\ul \textit{400}} &
  {\ul \textit{560}} &
  {\ul \textit{720}} &
  {\ul \textit{880}} &
  {\ul \textit{1000}} \\ \midrule
convSeq2Seq  \cite{Li18}         & 20.6 & 39.6 & 59.7 & 67.6 & 79.0 & 87.0 & 91.5 & 93.5 \\
LTD-10-10 \cite{Mao19LTD}            & 10.3 & 19.3 & 36.6 & 44.6 & 61.5 & 75.9 & 86.2 & 91.2 \\
LTD-10-25 \cite{Mao19LTD}            & 11.0 & 20.7 & 37.8 & 45.3 & 57.2 & 65.7 & 71.3 & 75.2 \\
DCT-RNN-GCN \cite{Mao20DCT}             & 11.3 & 20.7 & 35.7 & 42.0 & 51.7 & 58.6 & 63.4 & 67.2 \\ \midrule
\multicolumn{1}{|c|}{Ours} &
  \textbf{10.0} &
  \textbf{12.5} &
  \textbf{21.8} &
  \textbf{24.5} &
  \textbf{31.9} &
  \textbf{38.1} &
  \textbf{42.7} &
  \textbf{45.5} \\ \bottomrule
\end{tabular}}
\caption{Average MPJPE in mm over the \textit{BMLrub} test sequences of AMASS.
Our model outperforms the current best~\cite{Mao20DCT} by 32\% for 25-frame (1000 msec) predictions.
}
\label{tab:amass_tab}
\end{table}

\myparagraph{AMASS} Also in the case of AMASS, in Table~\ref{tab:amass_tab}, for short and long-term predictions of 3D coordinates, our model outperforms the state-of-the-art by 32\% on the longest time horizon (25-frame, 1000 msec).

\myparagraph{3DPW}
In Table~\ref{tab:tab_3dpw}, we test the generalizability of our model by training on AMASS and testing on 3DPW. Results are significantly beyond the state-of-the-art. For 2-frame predictions we reduce the error by 32\%, compared to the second best. For any other time horizon above 4 frames, we reduce the error by at least 43\%.

\begin{table}[h!]
\centering
\resizebox{0.49\textwidth}{!}{
\begin{tabular}{|c|cccccccc|}
\cline{2-9}
\multicolumn{1}{l|}{} & \multicolumn{8}{c|}{3DPW}                             \\ \hline
\multicolumn{1}{|c|}{\textit{msec}} &
  {\ul \textit{80}} &
  {\ul \textit{160}} &
  {\ul \textit{240}} &
  {\ul \textit{400}} &
  {\ul \textit{560}} &
  {\ul \textit{720}} &
  {\ul \textit{880}} &
  {\ul \textit{1000}} \\ \midrule
convSeq2Seq  \cite{Li18}         & 18.8 & 32.9 & 52.0 & 58.8 & 69.4 & 77.0 & 83.6 & 87.8 \\
LTD-10-10   \cite{Mao19LTD}          & 12.0 & 22.0 & 38.9 & 46.2 & 59.1 & 69.1 & 76.5 & 81.1 \\
LTD-10-25    \cite{Mao19LTD}         & 12.6 & 23.2 & 39.7 & 46.6 & 57.9 & 65.8 & 71.5 & 75.5 \\
DCT-RNN-GCN \cite{Mao20DCT}             & 12.6 & 23.1 & 39.0 & 45.4 & 56.0 & 63.6 & 69.7 & 73.7 \\ \midrule
\multicolumn{1}{|c|}{Ours} &
  \textbf{8.6} &
  \textbf{12.8} &
  \textbf{21.0} &
  \textbf{24.5} &
  \textbf{30.4} &
  \textbf{35.7} &
  \textbf{39.6} &
  \textbf{42.3} \\ \bottomrule
\end{tabular}}
\caption{Average MPJPE in mm, testing the generalizability on 3DPW of models trained on AMASS. Our model scores significantly beyond the state-of-the-art, i.e.\ it outperforms \cite{Mao20DCT}, on 4-25 frames (160-1000 msec) by at least 43\%.
}
\label{tab:tab_3dpw}
\end{table}

\myparagraph{Number of parameters} Table \ref{tab:ablation} (\textit{rightmost column}) compares the number of parameters of our model Vs.\ \cite{Mao20DCT}. Ours uses a fraction of parameters, $57.5k$ Vs $3.4M$, only 1.7\%.

\myparagraph{Qualitative evaluation}
We provide sample predictions (\textit{purple/green}) in Fig.~\ref{fig:seq_pred} on Human3.6M against ground truth sequences (\textit{gray/black}).
All predictions are long-term (25 frames) but we only display one every three frames for \emph{Discussion} and \emph{Posing}, to fit the illustrations into a row.
Results are in line with the long-term error statistics of Table~\ref{tab:long_mm}. The forecast \emph{Walking} is accurate, within 5.2 cm-accuracy in average at 25 frames (1 sec) and pictorially matching the ground truth. This shows how our model learns periodic motion well.
Predicted future poses are also relatively accurate for \emph{Discussion}, where the average error is 7.9 cm (cf.\ competing algorithms are nearly 12 cm). In this case, our model predicts well the mostly static pose of the discussing person, but the error is larger on the waving left hand.
Finally our model is producing larger errors on \emph{Posing} (10.6 cm in average), as it is a more challenging aperiodic action, which different people perform in different ways.

\begin{table}[!htbp]
\centering
\resizebox{0.49\textwidth}{!}{
\begin{tabular}{|c|cccccccc|c|}
\cline{2-9}
\multicolumn{1}{c|}{}     & \multicolumn{8}{c|}{Average}                                   \\ \hline
\multicolumn{1}{|c|}{\textit{msec}} &
  {\ul \textit{80}} &
  {\ul \textit{160}} &
  {\ul \textit{240}} &
  {\ul \textit{400}} &
  {\ul \textit{560}} &
  {\ul \textit{720}} &
  {\ul \textit{880}} &
  {\ul \textit{1000}} & Parameters \\ \midrule
DCT-RNN-GCN \cite{Mao20DCT}     & 10.4 &
  22.6 &
  47.1 &
  58.3 &  77.3 &
  91.8 &
  104.1 &
  112.1 & 3.4M \\  
 \midrule
Distinct $\mathcal{G}^s$, $\mathcal{G}^t$   & 28.9 & 26.4 & 40.2 & 48.7 & 58.7 & 66.9 & 75.2          & 79.9 & 59.8k \\
Full $\mathcal{G}^{st}$   & 11.9 & 19.4 & 34.1 & 40.8 & 53.1 & 65.6 & 75.1          & 82.5 & 222.9k \\
Separable $\mathcal{G}^{s-t}$ shared   & 11.3 & 19.4 & 34.7 & 40.5 & 52.5 & 62.1 & 69.2 & 76.9 & 36.4k\\
\multicolumn{1}{|c|}{Separable $\mathcal{G}^{s-t}$ (proposed)} &
  \textbf{10.1} &
  \textbf{17.1} &
  \textbf{33.1} &
  \textbf{38.3} &
  \textbf{50.8} &
  \textbf{60.1} &
  \textbf{68.9} &
  \textbf{75.6} & 57.5k\\ \bottomrule
\end{tabular}}
\caption{Average MPJPE error in mm on Human3.6M, comparing ablating variants of our model. See \ref{marker_abl_sec} for the detailed discussion. We also report here the number of parameters of all techniques, as well as of the current best algorithm~\cite{Mao20DCT}. Our proposed Separable $\mathcal{G}^{s-t}$ has only 1.7\% of the parameters of \cite{Mao20DCT}.
}
\label{tab:ablation}
\end{table}

\subsection{Ablation Study}
\label{marker_abl_sec}

Table~\ref{tab:ablation} illustrates the following ablative variants of our proposed STS-GCN encoding technique:

\myparagraph{Distinct graphs $\mathcal{G}^s$ and $\mathcal{G}^t$} This stands for separate GCNs for space and time, with separate adjacency and projection matrices, intertwined by an activation function. The variant underperforms our proposed model, which confirms the importance of spatio-temporal interaction within a single graph $\mathcal{G}^{st}$.
Interestingly the errors are much larger for the short-range (nearly 3x larger) than for the long-range (+6\% errors). We believe longer-term correlations may aid the variant.

\myparagraph{Full (non-separable) graph $\mathcal{G}^{st}$}
The variant adopts a full space-time adjacency matrix $A^{st}$. We observe a similar trend as for distinct graphs, i.e.\ worse performance with larger error increase (+18\%) for short-term predictions but better for long-term ones (+9\%). Notably the full graph model requires nearly 4x more parameters than our proposed one, cf.\ rightmost column in Table~\ref{tab:ablation}.

\myparagraph{Separable graph $\mathcal{G}^{s-t}$ shared across layers}
This only differs as it learns shared adjacency matrices across all layers, rather than layer-specific ones. Errors are comparable in the long-term (+2\%) but larger in the short-term (+12\%), against saving 37\% of the parameters.

\begin{figure}[h!]
\centering
{\includegraphics[width=.48\textwidth]{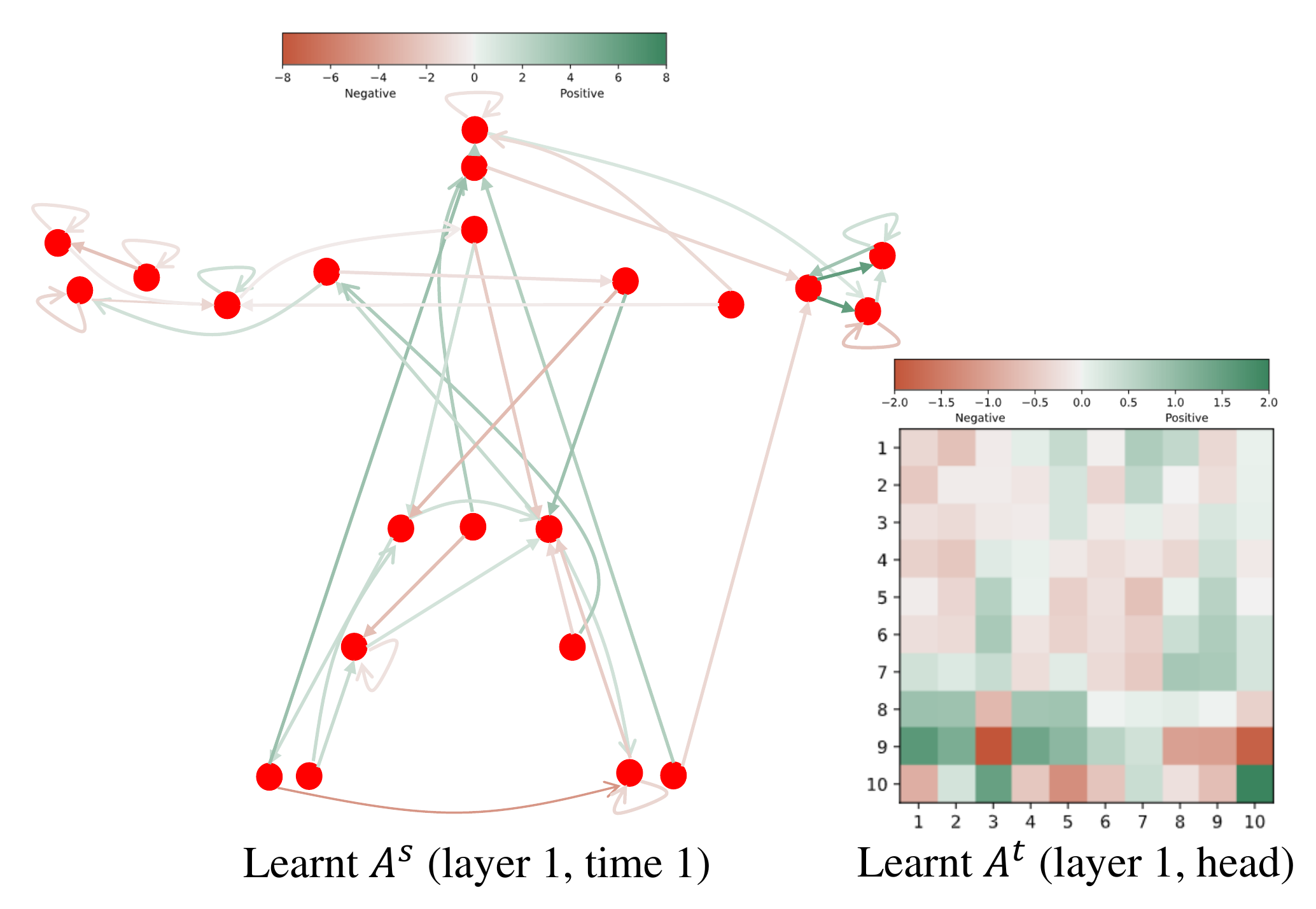} }
\caption{Sample spatial $A^s$ (\textit{left}) and temporal $A^t$ (\textit{right}) adjacency matrices, learnt on Human3.6M. 
(\textit{left}) Red dots represent the 22 human joints; learnt joint-joint relations mainly follow the kinematic tree, but additionally bring up long-term connections (e.g.\ foot-foot, head-foot) which support forecasting.
(\textit{right}) The learnt $A^t$ shows information flow from the earlier to the later observed frames, i.e.\ larger absolute values in the bottom-left part of the matrix. (Zoom in for details.) See Sec.~\ref{marker_abl_sec} for a discussion.
}
\label{fig:spatial_edges}
\end{figure}

\paragraph{Learnt separable space and time adjacency matrices}

In Fig.~\ref{fig:spatial_edges}, we illustrate two learnt adjacency matrices, upon training on Human3.6M. On the left, we represent a spatial adjacency matrix $A^s$, i.e.\ imagine the red dots positioned on the 22 keypoints of a frontally posing Vitruvian man. Learnt parameters are directed (as the learnt matrix $A^s$ is not symmetric) and signed edges (cf.\ Sec.~\ref{sec:sts-gcn-sub}), color-coded weights as in the legend. For clarity of illustration, we represent the two strongest connections for each keypoint. Note how most learnt connections follow the kinematic tree, which confirms the importance of the physical linkage. However additional strong connections also emerge, which bridge distant but motion-related joints, such as the two feet, the feet to the head, and the shoulders to the opposite hips, which intuitively interact for future pose prediction.\\
In Fig.~\ref{fig:spatial_edges} (\textit{right}), we represent a temporal adjacency matrix $A^t$, also asymmetric and signed.
It is noticeable the information flow from the earlier to the later observed frames. So the bottom-left side of the matrix shows larger absolute values. In particular most information is drawn to the last two frames (bottom two rows), corresponding to the 9th and 10th observed frames. Note also that the range of temporal relation coefficients $[-2,2]$ is smaller than the spatial $[-8,8]$, which privileges spatial information above the temporal when forecasting future poses.

\section{Conclusions}

We have proposed a novel Space-Time-Separable Graph Convolutional Network (STS-GCN) for pose forecasting. The single-graph framework favors the cross-talk of space and time, while bottleneck'ing the space-time interaction allows to better learn the fully-trainable joint-joint and time-time interactions. The model improves considerably on the state-of-the-art performance and but only requires a fractions of the parameters. These results further support the adoption of GCN and future research on it.

\section*{Acknowledgements}
The authors wish to acknowledge Panasonic for partially supporting this work and the project  of  the  Italian  Ministry  of  Education,  Universities and  Research  (MIUR)  ``Dipartimenti  di  Eccellenza 2018-2022''.

\newpage
{\small
\bibliographystyle{ieee_fullname}
\bibliography{egbib}
}

\end{document}